\documentclass[sigconf, nonacm]{acmart}

\usepackage{soul}

\AtBeginDocument{%
  }

\acmConference[]{}{}{}

\begin{document}

\title{Neuromorphic Bayesian Optimization in Lava}

\author{Shay Snyder}
\affiliation{%
  \institution{George Mason University}
  \streetaddress{4400 University Drive}
  \city{Fairfax}
  \state{Virginia}
  \country{USA}
  \postcode{22030}
}
\email{ssnyde9@gmu.edu}

\author{Sumedh R. Risbud}
\affiliation{%
  \institution{Intel Labs}
  \streetaddress{2200 Mission College Blvd}
  \city{Santa Clara}
  \state{California}
  \country{USA}
  \postcode{95054}
}
\email{sumedh.risbud@intel.com}

\author{Maryam Parsa}
\affiliation{%
  \institution{George Mason University}
  \streetaddress{4400 University Drive}
  \city{Fairfax}
  \state{Virginia}
  \country{USA}
  \postcode{22030}
}
\email{mparsa@gmu.edu}

\renewcommand{\shortauthors}{Shay Snyder, Sumedh Risbud, \& Maryam Parsa}

\begin{abstract}
  The ever-increasing demands of computationally expensive and high-dimensional problems require novel optimization methods to find near-optimal solutions in a reasonable amount of time. Bayesian Optimization (BO) stands as one of the best methodologies for learning the underlying relationships within multi-variate problems. This allows users to optimize time consuming and computationally expensive black-box functions in feasible time frames. Existing BO implementations use traditional von-Neumann architectures, in which data and memory are separate. In this work, we introduce Lava Bayesian Optimization (LavaBO) as a contribution to the open-source Lava Software Framework. LavaBO is the first step towards developing a BO system compatible with heterogeneous, fine-grained parallel, in-memory neuromorphic computing architectures (e.g., Intel's Loihi platform). We evaluate the algorithmic performance of the LavaBO system on multiple problems such as training state-of-the-art spiking neural network through back-propagation and evolutionary learning. Compared to traditional algorithms (such as grid and random search), we highlight the ability of LavaBO to explore the parameter search space with fewer expensive function evaluations, while discovering the optimal solutions.
\end{abstract}




\maketitle

\section{Introduction}
At the core of a vast array of problems lies a large function, evaluation of which is the most computationally expensive step: neural network design~\cite{parsa2021accurate}, transportation systems~\cite{sha2020applying}, graph neural networks~\cite{cong2022semi}, and evolutionary algorithms~\cite{parsa2021multi} are some examples. The common attribute in these problems is the computational complexity of the evaluation.
Computer scientists and mathematicians dedicate tremendous amounts of time and energy developing and optimizing a plethora of algorithms for finding optimal or near-optimal solutions to such problems. The algorithms are either rooted in the study of the theory of computation~\cite{snyman2005practical} or the development of statistical models of relationships between multiple variables~\cite{sun2019survey}.
The use of Bayesian Optimization (BO), based on Bayes' theorem, belongs to the latter class. Published in 1763, Bayes' theorem~\cite{bayes1763lii} revolutionized the field of conditional probability and probabilistic inference. It allows Bayesian systems to use their prior knowledge to construct probabilistic models of the world. There are many applications ranging from communication systems~\cite{middleton1960introduction}, medical diagnosis~\cite{kononenko1993inductive}, and quantitative finance~\cite{rachev2008bayesian} to hyperparameter optimization~\cite{parsa2019pabo, parsa2021multi, parsa2020bayesian}. 

In this work, we introduce Lava Bayesian Optimization (LavaBO), which adds support for BO in the open-source Lava Software Framework~\cite{lavaswfmk} and makes it available to the neuromorphic community.\footnote{Code available as a part of {\tt https://github.com/lava-nc/lava-optimization}} LavaBO is the first step towards developing a BO system compatible with fine-grained parallel neuromorphic computing architectures (e.g., Intel's Loihi platform~\cite{8259423}). Specifically, using the CPU-based implementation of LavaBO, we evaluate its algorithmic performance on multiple problems: the classical non-convex case of the Ackley function~\cite{ackley1987model}, a classification problem using an evolutionary algorithm (EONS~\cite{10.1145/3381755.3381758}) on the TENNLab framework~\cite{8573122}, and a deep spiking neural network performing classification problem on the NMNIST dataset~\cite{10.3389/fnins.2015.00437} using the SLAYER algorithm~\cite{tim_head_2018_1207017} from the Lava deep learning library~\cite{lava-dl}.

We compare our results with grid and random search algorithms to highlight LavaBO's ability to uncover optimal solutions with fewer numbers of expensive evaluations of the corresponding black-box functions. We conclude this article with a detailed discussion of the broader impact of this technology and our plan to implement it on Intel's Loihi 2 neuromorphic chip~\cite{loihi2techbrief}.



\begin{figure}[ht!]
    \centering
    \includegraphics[width=0.48\textwidth]{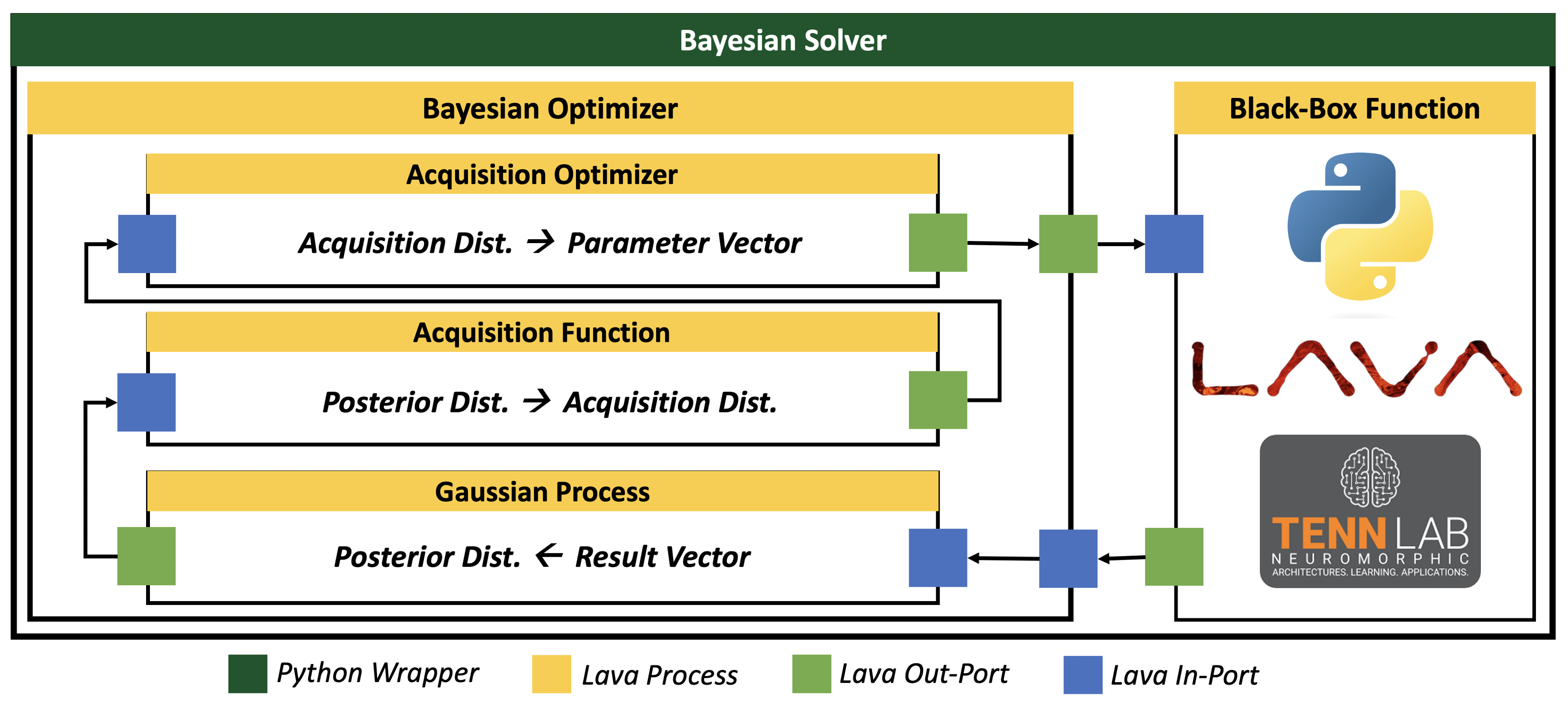}
    \caption{A flowchart presenting the programmatic architecture of the Lava Bayesian Optimization system.}
    \label{fig:arch}
\end{figure}




\section{Architecture of LavaBO within the Lava Framework}
Lava Software Framework is an open-source software framework, aimed at lowering the barrier for programming neuromorphic hardware and prototyping neuro-inspired algorithms~\cite{lavaswfmk}. It achieves this goal by providing the necessary tools and abstractions for creating applications that can run on different types of neuromorphic platforms, such as Intel’s Loihi chips. Lava also supports conventional CPUs and GPUs for prototyping and testing purposes. Lava aims to be a modular, composable, and extensible framework that allows researchers to integrate their ideas into a growing algorithms library and build complex neuromorphic applications. One such library built on top of the Lava primitives is the Lava Optimization library.
LavaBO is implemented with the same interface structure as the rest of the solvers within the Lava-Optimization library. This leads to a situation where the main interface between users and LavaBO is a wrapper class that initializes, connects, and executes all underlying Lava Processes\footnote{See {\tt http://lava-nc.org} for details about Lava concepts like a {\em Process}.}. Figure~\ref{fig:arch} shows a detailed look at the structure of LavaBO. The code and a tutorial on how to leverage this tool are accessible at \texttt{https://github.com/lava-nc/lava-optimization}.


\subsection*{Bayesian Solver} The BayesianSolver abstracts lower-level functionalities of the optimization process away and serves as the interface contract between users and developers. The system is highly configurable with multiple acquisition functions, acquisition optimizers, initial point generators, the number of initial points, and the random state. Once the desired parameters are determined and the BayesianSolver is initialized, the final remaining steps are creating a Lava Process and ProcessModel wrapper around the black-box function. The only requirements for these wrappers are that they must receive an array of parameters and return an array of parameters along with the resulting score.

\subsection*{Gaussian Regressor} 
The Gaussian-Regressor learns through prior knowledge and constructs a statistical model of unknown points and their resultant values. This system is implemented as a floating-point Gaussian regressor based on equation~\ref{eq:GR}
\begin{equation}
    f(x)\sim \mathbb{GP}(m(x),k(x,x')) \label{eq:GR}
\end{equation}
where $m(x)$ is a mean function over a real process $f(x)$ and $k(x, x')$ is a covariance function over the same $f(x)$~\cite{10.5555/1162254}.

\subsection*{Acquisition Function} 
We will explore and exploit the search space through the acquisition function.
The AcquisitionFunction defines and calculates a function on the posterior distribution from the GaussianRegressor that represents uncertainly across the space.
LavaBO supports three different acquisition functions: lower confidence bound~\cite{ROSS2017329}, negative expected improvement~\cite{agnihotri2020exploring}, and negative probability of improvement~\cite{agnihotri2020exploring}.

\subsection*{Acquisition Optimizer} The AcquisitionOptimizer determines how points are selected based on the acquisition distribution from the AcquisitionFunction. For example, one could calculate the acquisition function across all points in the space or intelligently or randomly select a subset to calculate. We currently support two methods: random sampling~\cite{acharya2013sampling} and inverse Hessian matrix estimation~\cite{WANG201941}.

\subsection*{Closing the loop} The process begins at the initial point sampler (IPS), where points are broadly sampled to construct the initial surrogate model of the space within the GaussianRegressor. The Acquisition-Distribution then calculates an acquisition distribution from the GaussianRegressor's posterior distribution that models uncertainty throughout the search space. The AcquisitionOptimizer uses the acquisition distribution to determine which point to select for evaluation. The selected point is transferred from the Bayesian-Optimizer to the user's black-box function where the given hyperparameter configuration will be evaluated. After evaluation, the black-box function transfers the result back to the BayesianOptimizer.

The BayesianOptimizer has new information about the underlying search space from the black-box process. Therefore, the next step is to transfer this information to the GaussianRegressor where it will incorporate this result into its statistical model of the space. The newfound posterior distribution from the GaussianRegressor is then sent to the AcquisitionFunction. Here, the posterior distribution is used to update acquisition distribution. Lastly, the acquisition distribution is sent to the AcquisitionOptimizer where the next point is determined.

\section{Results}
To highlight the capability of LavaBO, we chose three experiments for optimizing: 
\begin{enumerate}
    \item a standard two dimensional function, the Ackley function~\cite{ackley1987model}. 
    \item hyperparameters of an evolutionary algorithm for training a spiking neural network to classify the IRIS dataset~\cite{Dua:2019}.
    \item hyperparameters of a backpropagation-based learning algorithm, SLAYER from the Lava deep learning library~\cite{lava-dl}. 
\end{enumerate}
We use random search or grid search algorithms as a baseline performance measurement for all experiments. Table \ref{tab:resultssummary} summarizes the results from all experiments. In the subsections following the table, we discuss the details of each experiment.

\begin{table}[h!]
    \centering
    \resizebox{0.45\textwidth}{!}{%
    \begin{tabular}{|c|c|c|c|c|}
        \hline
        Experiment & Baseline algo. & \multicolumn{2}{|c|}{Function eval.s} & Gain \\
        & & Baseline & LavaBO & \\
        \hline
        Ackley function & Random search & 500 & 50 & 10X \\
        Evolutionary algo. & Grid search & 432 & 20 & 21X \\
        Deep SNN training$^\dagger$ & Random search & 28 & 8 & 3.5X\\
        \hline
    \end{tabular}
    }
    \caption{Summary of results from all three experiments detailed in this work. $^\dagger$In the case of training of deep SNNs, number of network evaluations to reach the same accuracy was counted. This iso-accuracy level is $\sim$30\% after 10 training epochs, as evident from Figure \ref{fig:slayer_nmnist_obs_box}.}
    \label{tab:resultssummary}
\end{table}




\subsection*{The Ackley function} 
\begin{figure}
    \centering
    \includegraphics[width=0.36\textwidth]{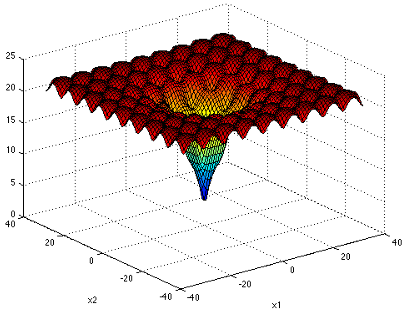}
    \caption{The Ackley function where $x_i \in [-32.768, 32.768]$ along with $a = 20$, $b = 0.2$, and $c = 2\pi$.}
    \label{fig:ackley_3d}
\end{figure}

Finding the global minimum of the Ackley function is a classical benchmark problem for non-convex global optimization. The equation for Ackley function is given as~\ref{eq:ackley}.
\begin{equation}
    f(x) = -a\exp{\left(-b\sqrt{\frac{1}{d}\sum_{i=1}^d x^2_i}\right)} - \exp{\left(\frac{1}{d}\sum_{i=1}^d cos\left(cx_i\right)\right)} + \left(a + e\right) \label{eq:ackley}
\end{equation}
where $a, b, c \in \mathbb{R}$ and $x_i \in [-32.768, 32.768]$.
Figure~\ref{fig:ackley_3d} presents a visualization of the Ackley function within the $x_i$ bounds where $a = 20$, $b = 0.2$, and $c = 2\pi$.

\begin{figure}
    \centering
    \includegraphics[width=0.4\textwidth]{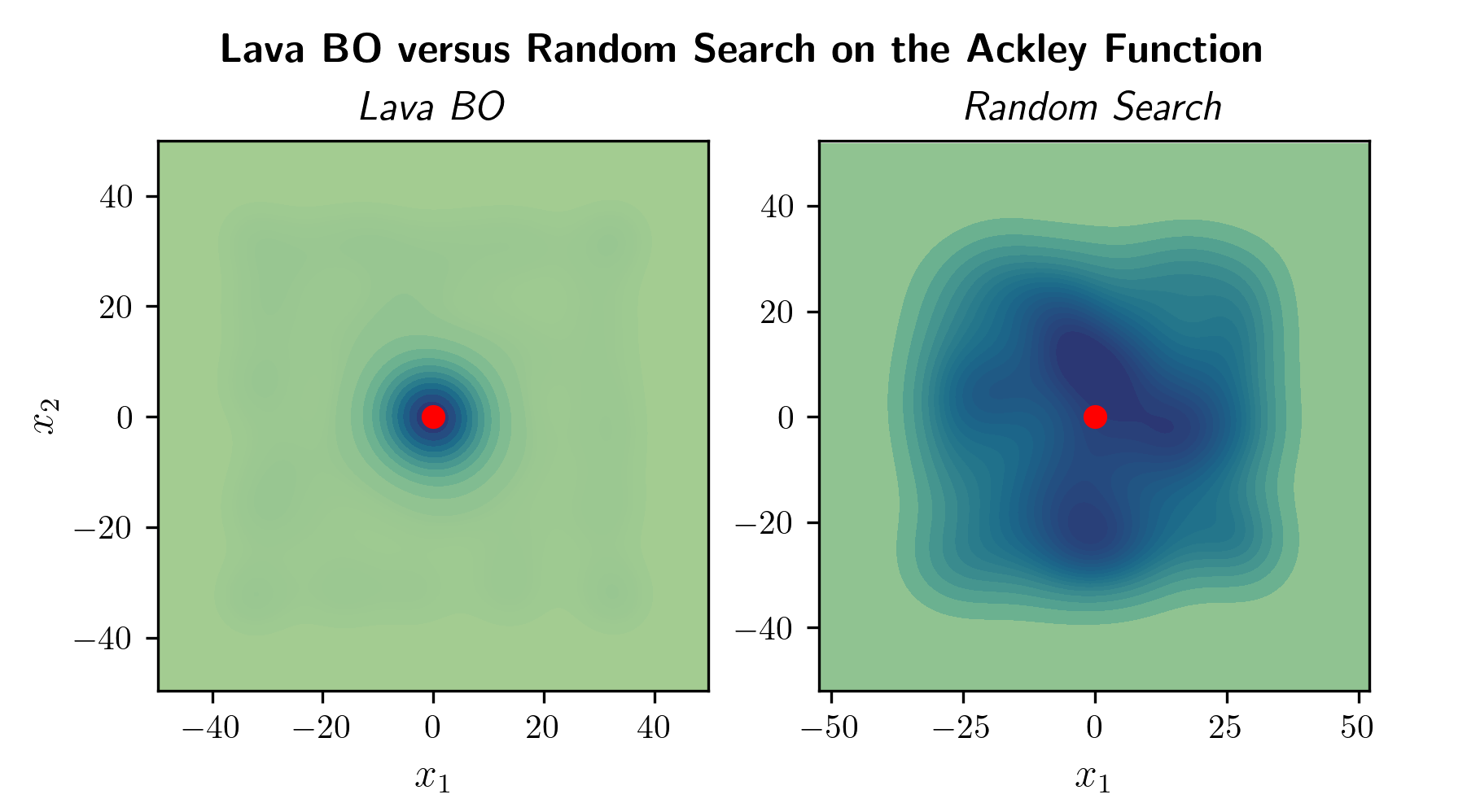}
    \caption{The probability distributions of observed points from Lava Bayesian Optimization and random search on the Ackley function. The red dots at $x_i = 0$ represent the optimum point within the search space.}
    \label{fig:ackley_bo_versus_grid}
\end{figure}

We conducted 10 LavaBO runs over the bounds of the Ackley function with each run consisting of randomly sampling 10 initial points from the search space before the surrogate model and acquisition function are used to intelligently observe 40 more points. A baseline random search was also conducted 10 times over the search space with each run consisting of randomly sampling and evaluating 50 points from the space. 
The probability distributions of the observed points for both algorithms are shown in Figure~\ref{fig:ackley_bo_versus_grid}.

There are stark performance differences between random search and LavaBO. As expected, the probability density function from the random search closely mirrors a uniform distribution. LavaBO's observations are tightly clustered around the minimum of the Ackley function, with a small percentage of the observations used in the fringes of the search space. 

\subsection*{Evolutionary Learning for IRIS Classification} 
\begin{figure}
    \centering
    \includegraphics[width=0.48\textwidth]{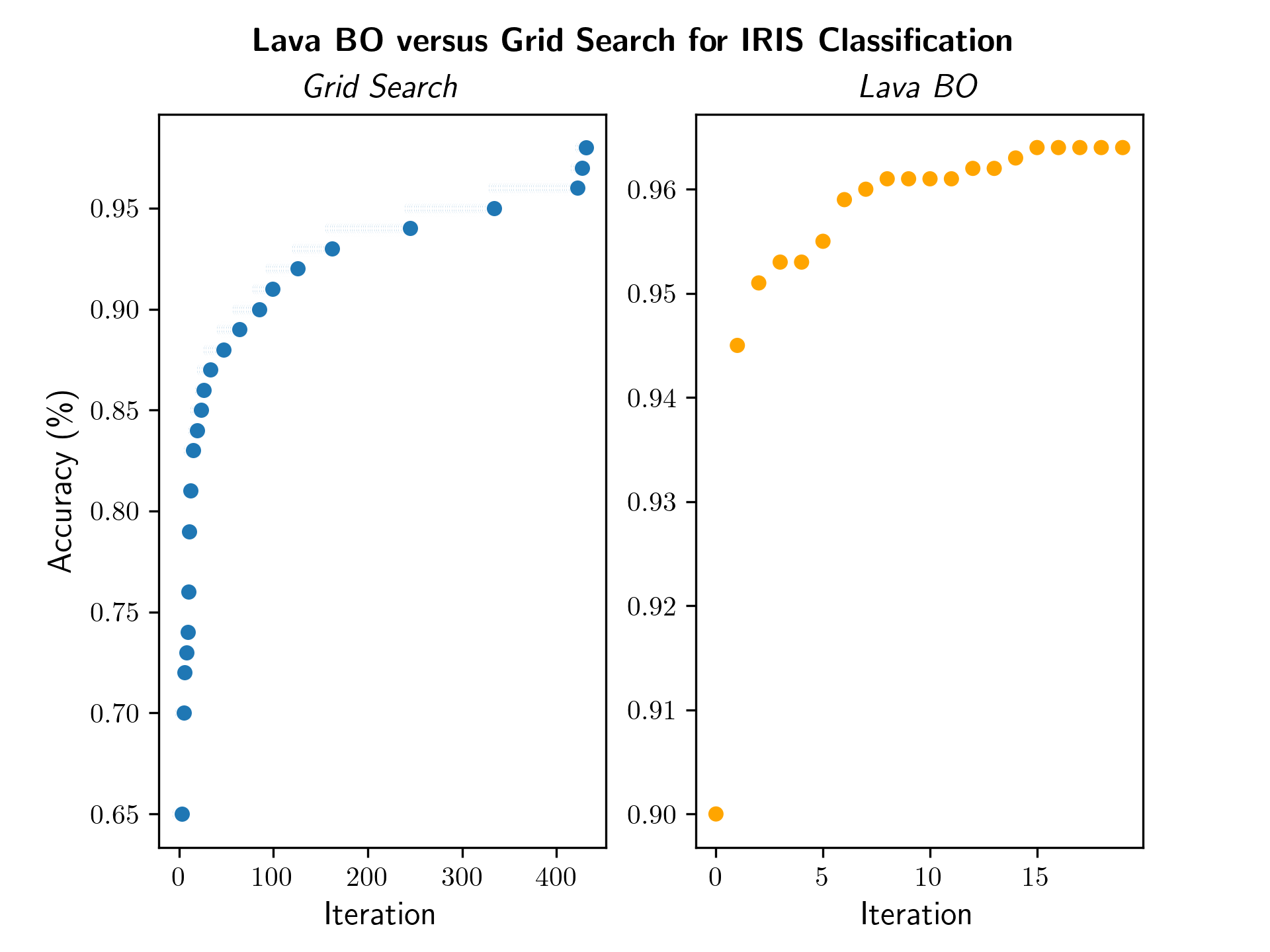}
    \caption{A comparison of LavaBO and grid search performance in optimization accuracy of IRIS classification using evolved spiking neural networks.}
    \label{fig:iris}
\end{figure}

In this experiment we compare performance of LavaBO and an exhaustive grid search on optimizing hyperparameters of an evolutionary-based learning algorithm (EONS)~\cite{schuman2016evolutionary} to solve a classification task on IRIS dataset~\cite{Dua:2019}. The details of the hyperparameters and search space can be found in Table~\ref{tab:iris_params}.
\begin{table}[]
    \begin{tabular}{|c|c|}
    \hline
    \textit{Parameter} & \textit{Options}   \\ \hline
    Crossover Rate          & 0.1, 0.3, 0.5, 0.7 \\
    Mutation Rate           & 0.1, 0.3, 0.5, 0.7 \\
    Num Mutations           & 1, 2, 3            \\
    Num Starting Edges      & 3, 5, 7            \\
    Num Starting Node       & 3, 5, 7            \\ \hline
    \end{tabular}
    \caption{The hyperparameter search space for the grid search and LavaBO on the evolutionary learning for IRIS classification problem. Search space size is 432.}
    \label{tab:iris_params}

\end{table}
We completed this task within the TENNLab framework~\cite{Dua:2019}. The grid search was performed over all 432 parameter combinations; however, the LavaBO was only ran for 20 iterations. We repeated the LavaBO process 10 times and averaged the final results to gain a broader understanding of the stability of the system. This is shown in Figure~\ref{fig:iris}. LavaBO was able to
learn the hyperparameter space and achieve $96.3\%$ accuracy after 16 Bayesian iterations.

\subsection*{SLAYER on Lava-DL for NMNIST Classification} 
In the third experiment, we demonstrate LavaBO's ability to find optimum hyperparameters of a deep spiking neural network with limited training epochs. We used the Lava deep learning library~\cite{10.3389/fnins.2015.00437} on the NMNIST dataset which aims to classify event-based handwritten digits. While the native training configuration requires 200 epochs, we limited the number of training epochs during the optimization process to 10 as many applications require models to converge very quickly. Therefore the goal of this experiment is to use LavaBO to find a hyperparameter configuration that achieves the highest validation accuracy during this limited training period.



The search space consists of parameters for voltage threshold of the spiking neurons (`threshold'), decay constants for current and voltage (`current decay' and `voltage decay', respectively), the time constant of the spike function derivative (`tau grad'), and the rate of change of the model parameters (the `learning rate'). More information on the specifics of each parameter and their ranges are shown in Table~\ref{tab:nmnist_params}.

\begin{table}[]
    \begin{tabular}{|c|c|c|c|}
        \hline
        \textit{Parameter} & \textit{Lower Bound} & \textit{Upper Bound} & \textit{Delta} \\ \hline
        Threshold          & 0.0                  & 5.0         & 0.125 \\
        Current Decay      & 0.0                  & 0.7         & 0.01  \\
        Voltage Decay      & 0.0                  & 0.7         & 0.01  \\
        Tau Grad           & 0.0                  & 0.7         & 0.01  \\
        Learning Rate      & $10^{-20}$         & 0.1        & 0.1   \\ \hline
    \end{tabular}
    \caption{The hyperparameter search space for the optimization problem with the Lava Deep Learning library, SLAYER, and NMNIST. All parameters have a series of discrete options between the min and max values. The individuals values are separated by the delta. This search space contains 137M parameter combinations.}
    \label{tab:nmnist_params}
\end{table}

\begin{figure}
    \centering
    \includegraphics[width=0.47\textwidth]{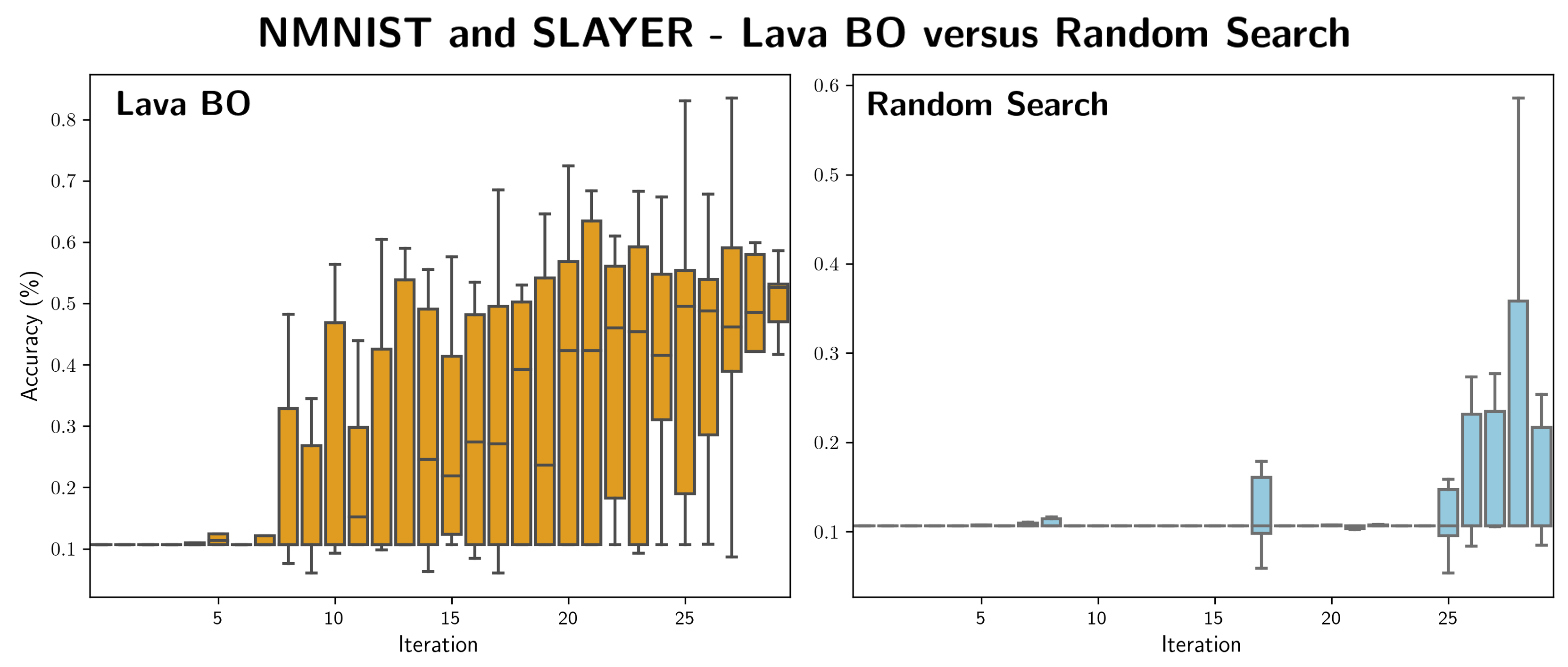}
    \caption{Side by side box plots of observation evaluations at every optimization iteration with Lava BO and random search. The ordering the box plots along the x-axis was sorted by their means. We see that the random search algorithm rarely ever evaluated hyperparameter configurations with greater than $25\%$ accuracy while Lava BO is consistently exploring points over $40\%$ accuracy while only training for $10$ epochs.}
    \label{fig:slayer_nmnist_obs_box}
\end{figure}

We ran random search and LavaBO on Lava-DL-SLAYER for NMNIST digit classification for 30 iterations and repeated each hyperparameter combination for three times. We only trained each hyperparameter combination for 10 epochs. 
Lava BO, on average, converges to a specific subset of the parameter search space that has over twice the accuracy of random search. This suggests that a subset of the space leads to more rapid convergence on the NMNIST digit classification task. Figure~\ref{fig:slayer_nmnist_obs_box} highlights this fact where Lava BO is constantly exploring new parameter combinations that have a much higher probability of performing well versus random search where the only high performing combinations are outliers.

\begin{figure}
    \centering
    \includegraphics[width=0.4\textwidth]{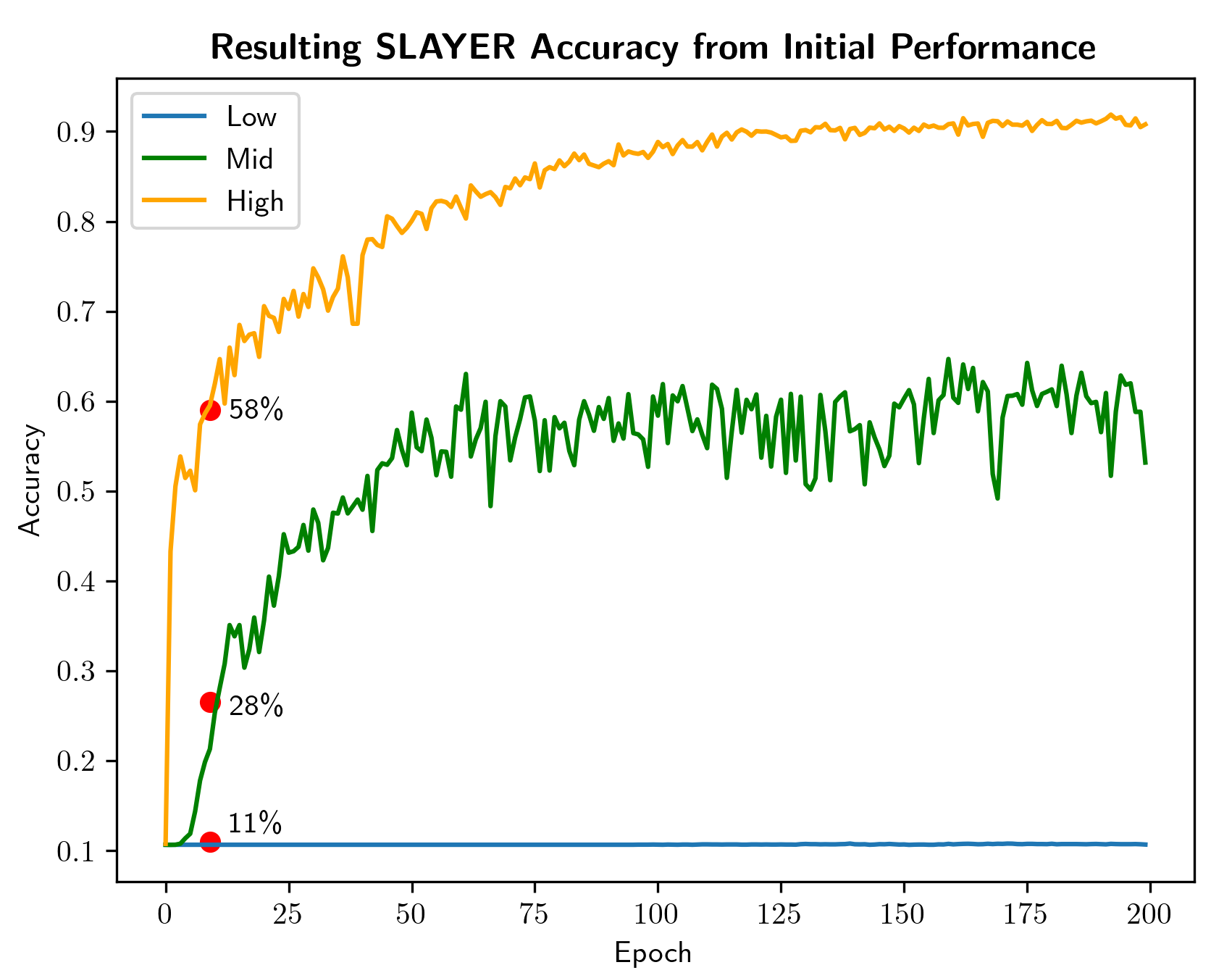}
    \caption{Accuracy plots for classifying NMNIST dataset for different hyperparameter combinations using SLAYER for 200 epochs. The three hyperparameter configurations are chosen based on low, medium, and high performing after training for 10 epochs.}
    \label{fig:slayer_initial2final}
\end{figure}

Evaluating the implications of our reduced number of training epochs, we chose 3 average configurations representing low, medium, and high performers with medians of $11\%$, $28\%$, and $58\%$ after 10 epochs, respectively. These categories and statistics were selected based on the distribution of results from Figure~\ref{fig:slayer_nmnist_obs_box}. For these parameter combinations we continued the training for 200 epochs. Figure~\ref{fig:slayer_initial2final} highlights the accuracy values for training for 200 epochs. For each combination, this figure also shows the original accuracy values after 10 epochs. The low performing model never improved its accuracy. The medium configuration represents a midpoint between the low and high categories, where the the final accuracy is between the upper and lower categories. Lastly, the high performance model achieved around $90\%$ accuracy after 200 epochs. 
We conclude that the highest performing models during the initial 10 epochs training have a higher likelihood of performing well at the end of full training sessions (200 epochs).

\section{Discussion}
In this work, we introduced the open-access Lava Bayesian Optimization (LavaBO) solver within the Lava Software Framework for neuromorphic computing. We outlined the architecture of the solver and compared its  algorithmic performance with traditional algorithms like random search, with the help of three problems: optimization of the Ackley function, hyperparameter tuning for evolutionary algorithms, and hyperparameter tuning for training of deep spiking neural networks. The results consistently show that LavaBO finds optimal hyperparameter combinations in significantly less iterations than random search.

As the natural next step, we are going to extend the LavaBO solver to be compatible with constraints imposed by the Loihi 2 neuromorphic chip. We hope that the extension will enable us to accelerate the execution of the solver using Loihi 2, while significantly reducing the power consumption. Such extension requires us to focus on two aspects: (a) Loihi 2 supports only fixed-point integer arithmetic with limited precision. As a result, we are investigating the effects of rounding on the accuracy of our solver, (b) additionally, for neural implementation of various components of LavaBO (e.g., the Gaussian Process regression), we are exploring the area of hyperdimensional (HD) computing using vector symbolic architectures~\cite{furlong2022fractional}. We hope this enables us to map all components of LavaBO efficiently on a neuromorphic substrate like Loihi 2.

\section*{Acknowledgment}
The work in this paper is supported by a gift from Intel Neuromorphic Research Community (INRC).

\balance

\bibliographystyle{ACM-Reference-Format}
\bibliography{main}

\end{document}